\documentclass[conference]{IEEEtran}
\IEEEoverridecommandlockouts
\usepackage{cite}
\usepackage{amsmath,amssymb,amsfonts,amsthm}
\usepackage{graphicx}
\usepackage{subcaption}
\usepackage{caption}
\usepackage{float}
\usepackage{textcomp}
\usepackage{xcolor}
\usepackage{booktabs}
\usepackage{bm}
\usepackage{algorithm,algorithmic}
\usepackage{nicefrac}
\usepackage{mathrsfs}
\usepackage{placeins}
\usepackage[colorlinks=true, linkcolor=blue, urlcolor=blue, citecolor=blue]{hyperref}
\usepackage{dsfont}

\newtheorem{theorem*}{Theorem}

\newtheorem{corollary*}{Corollary}

\newtheorem{proposition*}{Proposition}

\newtheorem{lemma*}{Lemma}

\newtheorem{definition}{Definition}

\def\BibTeX{{\rm B\kern-.05em{\sc i\kern-.025em b}\kern-.08em
    T\kern-.1667em\lower.7ex\hbox{E}\kern-.125emX}}

\begin{document}

\title{A Multiscale Geometric Method for Capturing Relational Topic Alignment\\

\thanks{The work in this paper was partially supported by grants from DOE NNSA Grant DE-NA0003921 and NSF grant CCF-2246213. PNNL-SA-213388. Code for reproducing results in this paper to be supported on GitHub: \href{https://github.com/ConradHougen/Multiscale-Topic-Manifold-Learning}{https://github.com/ConradHougen/Multiscale-Topic-Manifold-Learning}.}
}

\author{\IEEEauthorblockN{Conrad D. Hougen}
\IEEEauthorblockA{\textit{University of Michigan} \\
Ann Arbor, MI, USA \\
chougen@umich.edu}
\and
\IEEEauthorblockN{Karl T. Pazdernik}
\IEEEauthorblockA{\textit{Pacific Northwest National Laboratory} \\
Richland, WA, USA \\
karl.pazdernik@pnnl.gov}
\and
\IEEEauthorblockN{Alfred O. Hero}
\IEEEauthorblockA{\textit{University of Michigan}\\
Ann Arbor, MI, USA \\
hero@eecs.umich.edu}
}

\maketitle
\begin{abstract}
Interpretable topic modeling is essential for tracking how research interests evolve within co-author communities. In scientific corpora, where novelty is prized, identifying underrepresented niche topics is particularly important. However, contemporary models built from dense transformer embeddings tend to miss rare topics and therefore also fail to capture smooth temporal alignment. We propose a geometric method that integrates multimodal text and co-author network data, using Hellinger distances and Ward’s linkage to construct a hierarchical topic dendrogram. This approach captures both local and global structure, supporting multiscale learning across semantic and temporal dimensions. Our method effectively identifies rare-topic structure and visualizes smooth topic drift over time. Experiments highlight the strength of interpretable bag-of-words models when paired with principled geometric alignment.

\end{abstract}


\section{Introduction}\label{sec:introduction}
Topic modeling enables automated discovery of hidden semantic structure within documents. Traditional topic models operate on bag-of-words representations, using linear algebraic or probabilistic methods, like latent semantic analysis (LSA), non-negative matrix factorization (NMF), probabilistic LSA (pLSA), and latent Dirichlet allocation (LDA) \cite{Deerwester1990LSA, Lee1999NMF, Hofmann1999PLSA, blei2003latent}. These models use raw word counts or term-frequency inverse document frequency (TF-IDF) inputs \cite{SparckJones1972}. Word2Vec \cite{Mikolov2013word2vec} motivated usage of neural networks to translate words and phrases into dense vector embeddings. A similar approach is now used by large language models (LLMs) including BERT \cite{Devlin2019BERT}, RoBERTa \cite{liu2019roberta}, and GPT \cite{radford2018gpt}. Derivative topic models, like the contextualized topic model (CTM) \cite{bianchi2021_contextualized_topic_model}, BERTopic \cite{grootendorst2022bertopic}, GPTopic \cite{bianchi2023gptopic}, and TopicGPT \cite{topicgpt2023}, all use transformer embeddings.

While useful, these models operate on text alone, overlooking relational information within document networks. Relational topic models (RTMs) fill that gap \cite{wang2020deep, wang2020learning, wang2021layer, wang2024weibull, cai2011graph}. Citation networks, which are relatively dense and stable over time, are often used to relate documents \cite{giles1998citeseer}. In contrast, co-author networks are noisy, with local clique-structures and global sparsity. This complicates the integration of co-authorship data into RTMs. Prior works validate RTMs with link prediction benchmarks, but this is challenging in co-author networks.

Incorporating the temporal dimension into dynamic document networks introduces additional complexity. This is particularly true when the goal is to simultaneously align topics across time and across document networks. Few have attempted to tackle this joint challenge. Existing methods do not capture the time-evolving nature of author interests \cite{RosenZvi2012AuthorTopic}. Additionally, author-topic drift depends on the granularity at which topics are defined, motivating the development of a multiscale learning approach.

To address these challenges, we introduce Multiscale Topic Manifold Learning (MSTML), an ensemble model that combines temporal alignment \cite{wang2021geometry} with hierarchical learning \cite{clauset2008hierarchical} to capture multiscale structure in topics and co-author networks. MSTML integrates text and network data through an interpretable probabilistic model, aligning topics using Hellinger distances \cite{Hellinger1909} and Ward’s linkage \cite{ward1963hierarchical}, and visualizing them with Potential of Heat-diffusion for Affinity-based Trajectory Embedding (PHATE) embeddings \cite{Moon2019PHATE}. Unlike prior methods \cite{Hwang2017LDACoAuthorRecommender, Zhang2022DiffusionNetworkAnalytics}, MSTML relies only on texts and author lists, without external metadata. Additional method details and experiments can be found in \cite{hougen2025mstml, Hougen2025dissertation}.



\section{Background}\label{sec:background}
\subsection{Topic Modeling}
Topic modeling aims to uncover thematic structures in document corpora and must manage high-dimensional, noisy data. Large vocabularies are computationally costly \cite{Manning2008}, so preprocessing (stop word removal, stemming, lemmatization) and frequency-based filtering are applied to reduce size. Discarding single-occurrence terms is safe, while retaining moderate and high-frequency terms preserves topic granularity and document discriminative power \cite{lu2017vocabulary, Maier2020DocSampling}. For evaluating topic models, topic coherence metrics, such as $C_\text{V}$, $C_\text{UCI}$, and $C_\text{NPMI}$ \cite{Roder2015Coherence, newman2010automatic}, are popular, and they are part of our assessment. Topic coherence quantifies semantic similarity among top-$N$ words per topic embedding vector learned by a topic model, based on term co-occurrence statistics.

LDA is a generative probabilistic topic model and one of the most commonly-cited techniques \cite{blei2003latent}. In LDA, each word is modeled as an outcome of a generative process which randomly selects a topic $z$, then a word $w$. Topic $z$ is sampled according to multinomial distribution $\bm{\theta^{(j)}}$. $w$ is then sampled from the vocabulary, according to the multinomial distribution $\bm{\phi^{(z)}}$. This repeats for all $N_W^{(j)}$ words in document $j$, and for all documents $j\in\{1,\dots, N_D\}$. LDA uses Bayesian inference to learn the vectors of interest, $\{\bm{\theta^{(j)}}\}_{j=1}^{N_D}$ and $\{\bm{\phi^{(k)}}\}_{k=1}^{K}$.

Term relevancy (\ref{eqn:relevancy}) balances term frequency within a topic against exclusivity to that topic. We use term relevancy scores from an auxiliary model \cite{sievert-shirley-2014-ldavis} to rank and filter terms, which we found to improve model performance. $P(w \mid k)$ is the probability of term $w$ given the topic $k$, and $P(w)$ is the marginal probability of the term $w$ across the entire corpus. $\lambda\in(0,1)$ is a weight hyperparameter.
\vspace{-0.1in}
\begin{equation}\label{eqn:relevancy}
    r(w, k \mid \lambda) = \lambda \log P(w \mid k) + (1 - \lambda)\log \left( \frac{P(w \mid k)}{P(w)} \right)  
\end{equation}
\vspace{-0.2in} 

\subsection{Manifold Learning and Information Geometry}
Manifold learning is a nonlinear approach to reducing data dimensionality. Unlike linear methods like principal component analysis (PCA) \cite{Pearson1901PCA}, manifolds are assumed to be globally non-linear but locally linear. Algorithms include locally linear embedding (LLE), t-distributed stochastic neighbor embedding (t-SNE) and multi-dimensional scaling (MDS) \cite{Roweis2000LLE, vanDerMaaten2008TSNE, Kruskal1964MDS}. In this work, we use potential of heat-diffusion for affinity-based trajectory embedding (PHATE) \cite{Moon2019PHATE}, which handles time-evolving data effectively using density-adaptive diffusion to preserve local and global distances.

Preserving distances across multiple scales is also central to information geometry, which applies differential geometric tools to probability distributions. The $L_2$ distance is ineffective for comparing multinomial document-topic and topic-word distributions, such as $\bm{\theta}$ or $\bm{\phi}$ from LDA \cite{Nielsen2020Intro, Sun2014InfoGeo, wang2021geometry}. Instead, information geometry exploits the Fisher information metric, which defines a Riemannian structure on statistical manifolds. In practice, the Hellinger metric may be used as a computationally-simple approximation of the Fisher information. For multinomial vectors $\bm{p}$ and $\bm{q}$, the Hellinger metric is defined according to (\ref{eqn:hellinger}). The Hellinger distance is bounded between $0$ and $1$ and symmetric between $\bm{p}$ and $\bm{q}$.
\begin{equation}\label{eqn:hellinger}
    H(\bm{p}, \bm{q}) = \frac{1}{\sqrt{2}} \sqrt{\sum_{i=1}^{n} \left( \sqrt{\bm{p}_i} - \sqrt{\bm{q}_i} \right)^2 }
\end{equation}

For assessing longitudinal topic models, we define temporal topic alignment and manifold continuity metrics. Topic Neighborhood Overlap (TNO (\ref{eqn:tno})) measures the fraction of a topic’s $k$-nearest neighbors within a temporal window. Exponential Temporal Spectral Gap (ETSG (\ref{eqn:etsg})) measures algebraic connectivity of the $k$-NN graph of topic vectors after re-weighting edges by temporal distance, using an exponentially decaying kernel function. Another smoothness indicator is the number of components (NComp) in the $k$-NN topic graph.

\begin{definition}[Topic Neighborhood Overlap]
Let $\{\bm{\phi}_i\}_{i=1}^n$ be topic vectors with time labels $t_i \in \mathbb{Z}_{\geq 0}$. 
For each $\bm{\phi}_i$, let $\mathcal{N}_k(i)$ be its $k$-nearest vectors. 
For window size $w \geq 0$,
\begin{equation}\label{eqn:tno}
\mathrm{TNO}(w) = \frac{1}{n} \sum_{i=1}^n 
\frac{1}{k} \sum_{j \in \mathcal{N}_k(i)} 
\mathds{1}\left(|t_j - t_i| \leq w\right).
\end{equation}
\end{definition}
\begin{definition}[Exponential Temporal Spectral Gap]
Given a $k$-NN graph $G=(V, E),$ of topics with time labels $t_i \in \mathbb{Z}_{\geq 0}$, and $\tau > 0,$ $\epsilon > 0,$ assign edge weights
\[
w_{ij}^{(\tau)} = 
\begin{cases}
\exp\!\left(-\tfrac{|t_i - t_j|}{\tau}\right), & (i,j) \in E, \\[6pt]
\varepsilon, & (i,j) \notin E, \; i \neq j, \\[6pt]
0, & i = j,
\end{cases}
\]
Let ${W}^{(\tau)}$ be the weighted adjacency, $D^{(\tau)}$ the degree matrix, and 
$\mathscr{L}^{(\tau)} = I - (D^{(\tau)})^{-1/2} W^{(\tau)} (D^{(\tau)})^{-1/2}$ the normalized Laplacian, with $0=\lambda_1(\mathscr{L}^{(\tau)})\leq\cdots\leq \lambda_n(\mathscr{L}^{(\tau)})$. Then
\begin{equation}\label{eqn:etsg}
\mathrm{ETSG}(\tau) = \lambda_2(\mathscr{L}^{(\tau)}).
\end{equation}
\end{definition}

\subsection{Co-Author Networks}
A co-author network is defined by observable author lists associated with documents. When multiple authors share approximately the same name, author disambiguation is required, using fuzzy matching or additional meta-data \cite{Hwang2017LDACoAuthorRecommender}. A co-author network evolves over time, with each time window producing a snapshot of collaboration structure.
\begin{definition}[Co-Author Network]\label{def:coauth_net}
A \textbf{co-author network} is a graph \( \mathscr{G} = (\mathscr{V}, \mathscr{E}) \), where \( \mathscr{V} \) is the set of authors and \( \mathscr{E} \subseteq \mathscr{V} \times \mathscr{V} \) contains an edge \( (u, v) \) if authors \( u \) and \( v \) co-authored at least one document in the \textbf{active} document set. The graph is typically undirected, so \( (u, v) \in \mathscr{E} \) implies \( (v, u) \in \mathscr{E} \). The degree \( \deg(v) \) of a vertex is the number of co-authors of author \( v \).
\end{definition}

\section{Methods}\label{sec:methods}
\subsection{General Algorithm and Framework Diagram}
\begin{figure}[ht]
    \centering
    \includegraphics[width=.8\linewidth]{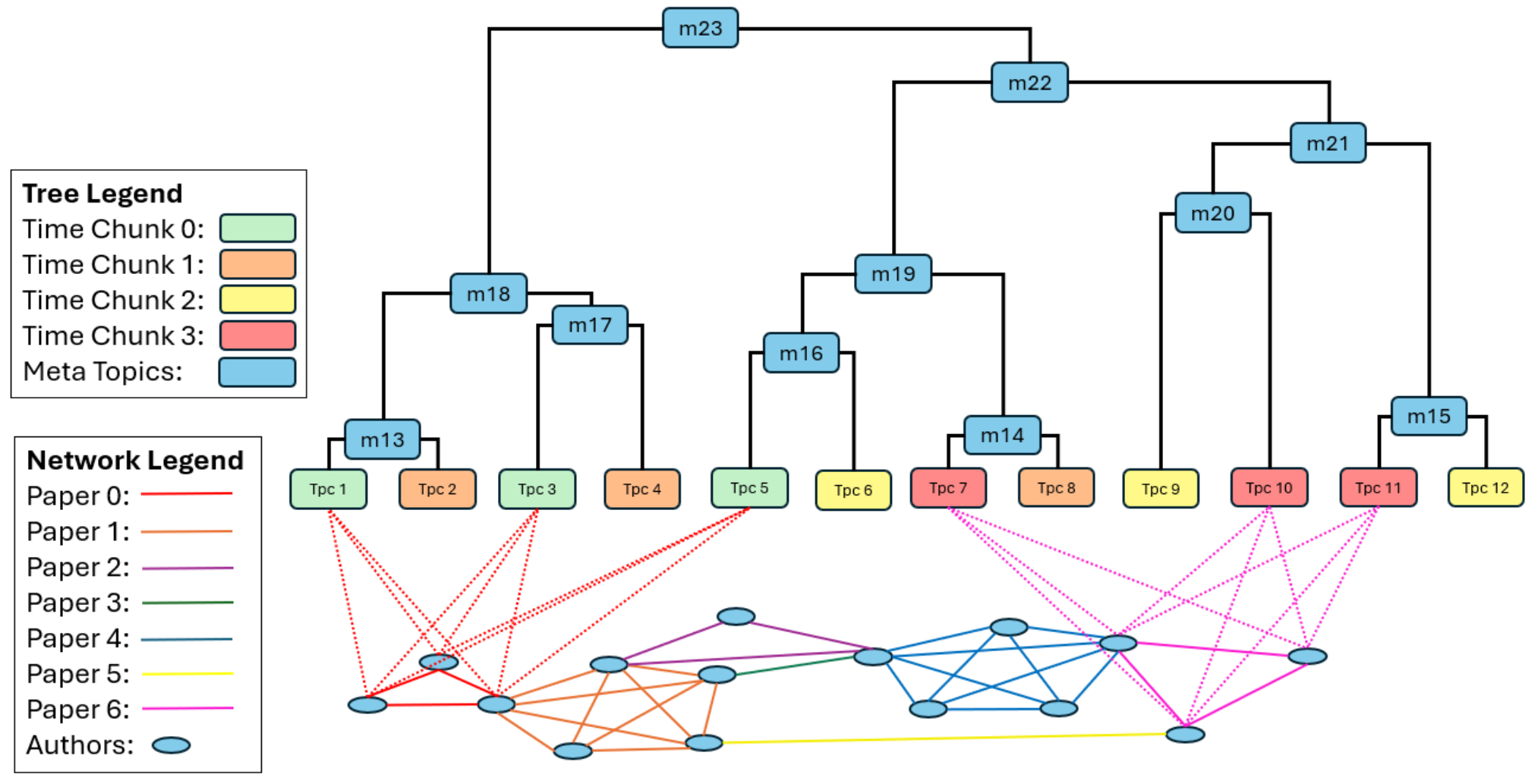}
    \caption{The topic space dendrogram (hierarchical tree, top) links chunk topics (multi-colored rectangles), meta-topics (m13-m23), and the co-author network (network, below). Several example links (red, pink) between the dendrogram and co-author network are included. These links represent author distributions over multiple chunk topic leaf nodes.} 
    \label{fig:tpc_dendro_model_combined}
\end{figure}

MSTML integrates topic manifold learning \cite{wang2021geometry} with hierarchical network models \cite{clauset2008hierarchical}.  The core method fits a dendrogram, parameterized by internal node probabilities, \(\{\mathscr{D}; \{p_m\}\}\), to a co-author network \(\mathscr{G}\). The authors (nodes) in $\mathscr{G}$ are represented by embeddings derived from an LDA topic model ensemble. LDA-derived topic vectors \(\{\bm{\phi^{(k)}}\}\) are mapped to dendrogram leaf nodes. These topic vectors reside in the probability simplex, \(\Delta^{\nu-1} \subset \mathbb{R}^\nu\), where \(\nu\) is the vocabulary size. The MCMC process of the HRG model, which learns the dendrogram topology, is replaced by agglomeratve clustering, using the Hellinger distance (\ref{eqn:hellinger}). Figure \ref{fig:tpc_dendro_model_combined} illustrates an example topic dendrogram model.

Instead of one LDA model, MSTML employs an ensemble learning approach. The corpus is split into uniform time chunks, with LDA applied to each sub-corpus \(C_1, C_2,\dots, C_T\) independently. Temporal smoothing is applied to improve continuity in the topic geometry \cite{wang2021geometry, hougen2025mstml, Hougen2025dissertation}. Each LDA model uses uniform Dirichlet priors, $\bm{\alpha}=\bm{1}, \bm{\beta}=\bm{1}$. A key challenge in LDA is selecting the number of topics, $K$, which is based on: avoiding topic duplication, high isolation between topics, and repeatability \cite{Gan2021LDA}. We have chosen to scale $K$ as an affine function of the number of documents per time chunk to balance topic continuity against discovery of emerging topics \cite{hougen2025mstml, Hougen2025dissertation}. Chunk topics are clustered into meta-topics using agglomerative clustering with Ward's linkage \cite{ward1963hierarchical}. The smooth topic manifold allows for interpretable traversal.

The ensemble approach mitigates overfitting, improving the odds of capturing niche topics that could be diluted by a single, global topic model. However, a global LDA model is uniquely used in MSTML for vocabulary filtering. We found that using global term relevancy (\ref{eqn:relevancy}) to rank and filter terms improves the interpretable quality of topic word clouds, with $\lambda\approx0.4$, adaptive based on the data \cite{hougen2025mstml, Hougen2025dissertation}. Additionally, the learned topic manifold appears more smooth after term relevancy filtering, with better topic alignment and clustering.

\subsection{Dendrogram Learning}
The MSTML topic dendrogram provides a multi-scale representation of topic relationships and links together text and co-author network data. \(\{\mathscr{D}; \{p_m\}\}\) is constructed in two stages. In the first, tree topology is determined using agglomerative hierarchical clustering of topic vectors \(\{\bm{\phi^{(k)}}\}\). Second, probabilities \(\{p_m\}\) are assigned to the internal nodes.

For the topology of $\mathscr{D}$, LDA-derived topic vectors \(\{\bm{\phi^{(k)}}\}\) are treated as multinomial distributions over the vocabulary \(\mathcal{V}\). A \(k\)-nearest neighbors graph, \(\mathcal{X}\), is constructed to capture meso-scale structures among topics to guide agglomerative clustering in alignment with PHATE embeddings. For large topic sets, FAISS is used to approximate the \(k\)-nearest neighbors graph \cite{Johnson2017FAISS}. Clustering proceeds by iteratively merging topic clusters, where each merge produces an internal node \( m \) with a height \( h_m \), representing inter-cluster dissimilarity. These heights \( h_m \) increase up the tree and may exceed the Hellinger distance bounds of \([0, 1]\) when using Ward's linkage. For consistent interpretations with various linkages and distances, heights are re-normalized to the \([0, 1]\) interval.

To compute the internal node probabilities \(\{p_m\}\), MSTML treats \(\mathscr{E}_m\), \(L_m\), and \(R_m\) as random variables based on author-topic distributions \(\{\bm{\psi^{(u)}}\}\), which are derived as weighted averages over their associated document-topic vectors. Each document-topic vector is weighted inversely by the number of authors on the document. The author-topic distributions map authors to chunk topics, which then map directly to dendrogram leaf nodes. Following \cite{clauset2008hierarchical}, the MLE estimator for internal node probabilities is defined as \(\hat{p}_m = \frac{\mathscr{E}_m}{L_m R_m}\), approximated here as \(\hat{p}_m \approx \frac{\mathbb{E}[\mathscr{E}_m]}{\mathbb{E}[L_m]\mathbb{E}[R_m]}\) under the assumption of weak correlation between \(\mathscr{E}_m\), \(L_m\), and \(R_m\).

Critically, this approach avoids the computational complexity of structure learning using MCMC, instead leveraging topic manifold clustering to estimate the dendrogram topology in one shot. Temporal topic trends and multi-scale author-topic relationships are captured. Internal node probabilities can be robustly estimated without concern for optimal structure estimation \cite{clauset2008hierarchical}. Probabilities \(\{p_m\}\) are computed as the ratio of the number of expected edges to possible edges, conditioned on LDA-learned topic distributions (\ref{eqn:p_m_mle}). $\hat{L}_m$, $\hat{R}_m$, and $\hat{\mathscr{E}}_m$ are computed as the expected numbers of authors in the left and right subtrees, and the number of edges between these subtrees, respectively \cite{Hougen2025dissertation}.
\begin{equation}
    \label{eqn:p_m_mle}
    p_m \triangleq \frac{\hat{\mathscr{E}}_m}{\hat{L}_m\hat{R}_m}, \forall m\in\mathscr{D}
\end{equation}


\section{Results}\label{sec:results}
This section shows experimental results using an arXiv corpus, collated by Cornell University and hosted on Kaggle (\href{https://www.kaggle.com/datasets/Cornell-University/arxiv}{https://www.kaggle.com/datasets/Cornell-University/arxiv}). The arxiv-stat-ml corpus is an extraction of 194,035 self-labeled document abstracts from statistics (stat) and machine learning (cs.LG) categories. These abstracts are supported on a vocabulary of 136,113 terms which is eventually filtered down to 8,465 terms. Code was written in Python and Cython.

\subsection{Visualizations and Multimodal Analysis}
MSTML visualizes the topic manifold, uncovering smooth topic alignment across time. The topic dendrogram can be cut at various heights. Using default MSTML parameters, a cut height of $h=0.55$ results in 9 meta topic clusters. These meta topic clusters can be identified by word clouds or by ranking top contributing documents. The 9 meta topics were mapped back to the top 30 contributing documents, using inferred probability mass. Topics were auto-labeled with ChatGPT: \{1: RL/Robotics, 2: Graph Learning, 3: Multimodal LMs, 4: Vision/DL, 5: Applied ML, 6: Bayesian Methods, 7: Opt/Bandits, 8: Classical ML, 9: Causal Inference\}.

\begin{figure}[htbp!]
    \centering
    
    \begin{subfigure}{0.7\linewidth}
        \centering
        \includegraphics[width=\linewidth]{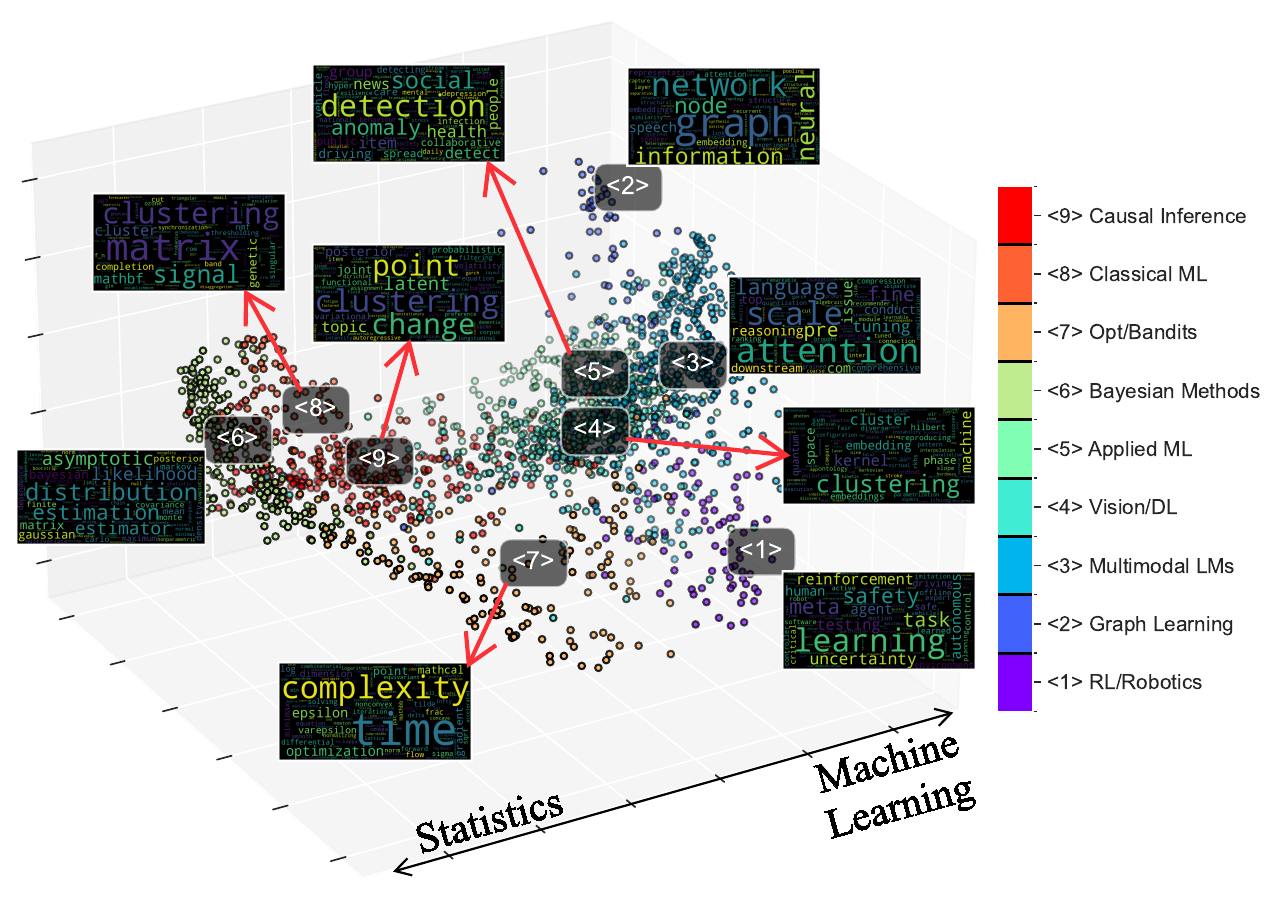}
        \label{fig:phate_9topics}
    \end{subfigure}
    
    \vspace{-0.1in} 
    
    \begin{subfigure}{0.48\linewidth}
        \centering
        \includegraphics[width=\linewidth]{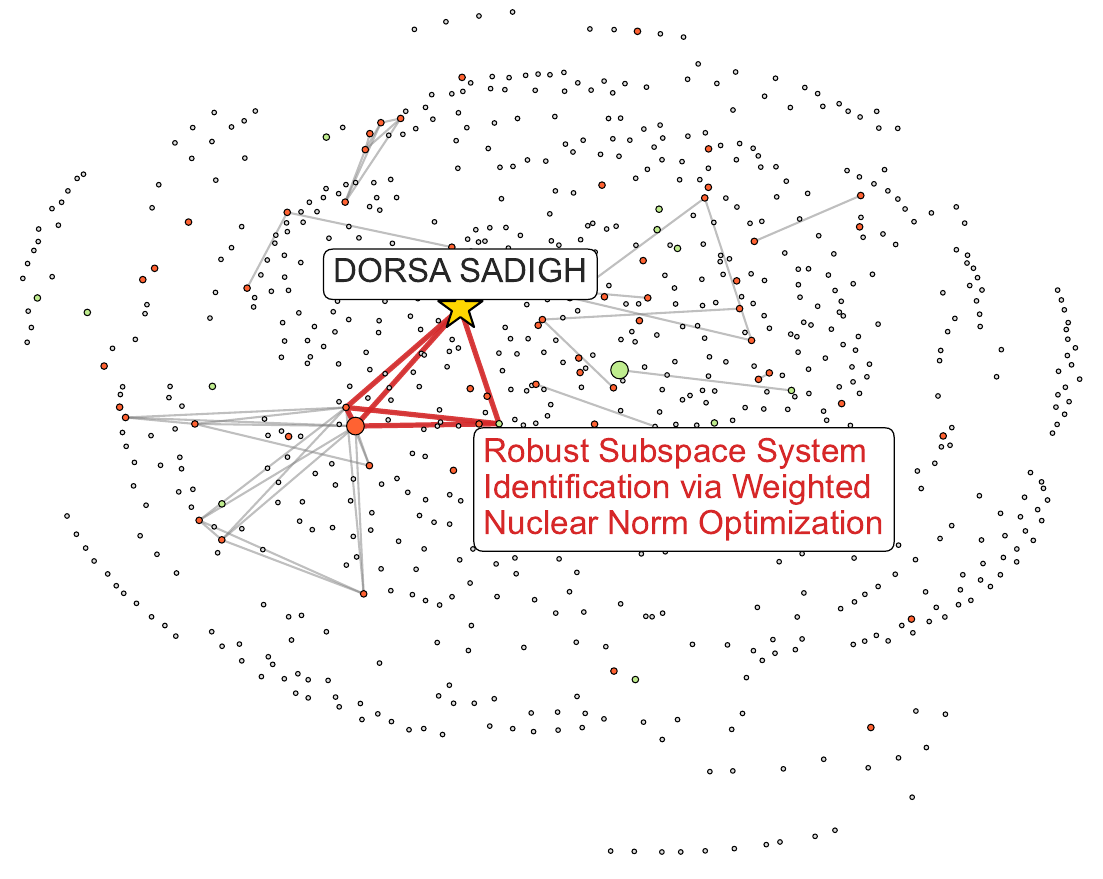}
    \end{subfigure}\hfill
    \begin{subfigure}{0.48\linewidth}
        \centering
        \includegraphics[width=\linewidth]{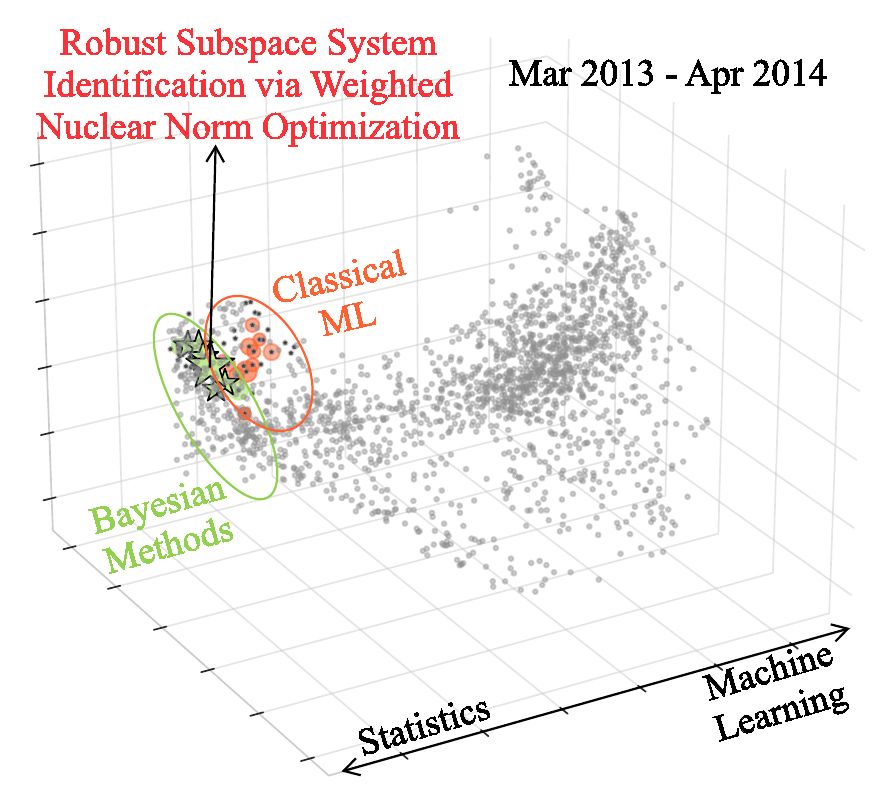}
    \end{subfigure}

    \begin{subfigure}{0.48\linewidth}
        \centering
        \includegraphics[width=\linewidth]{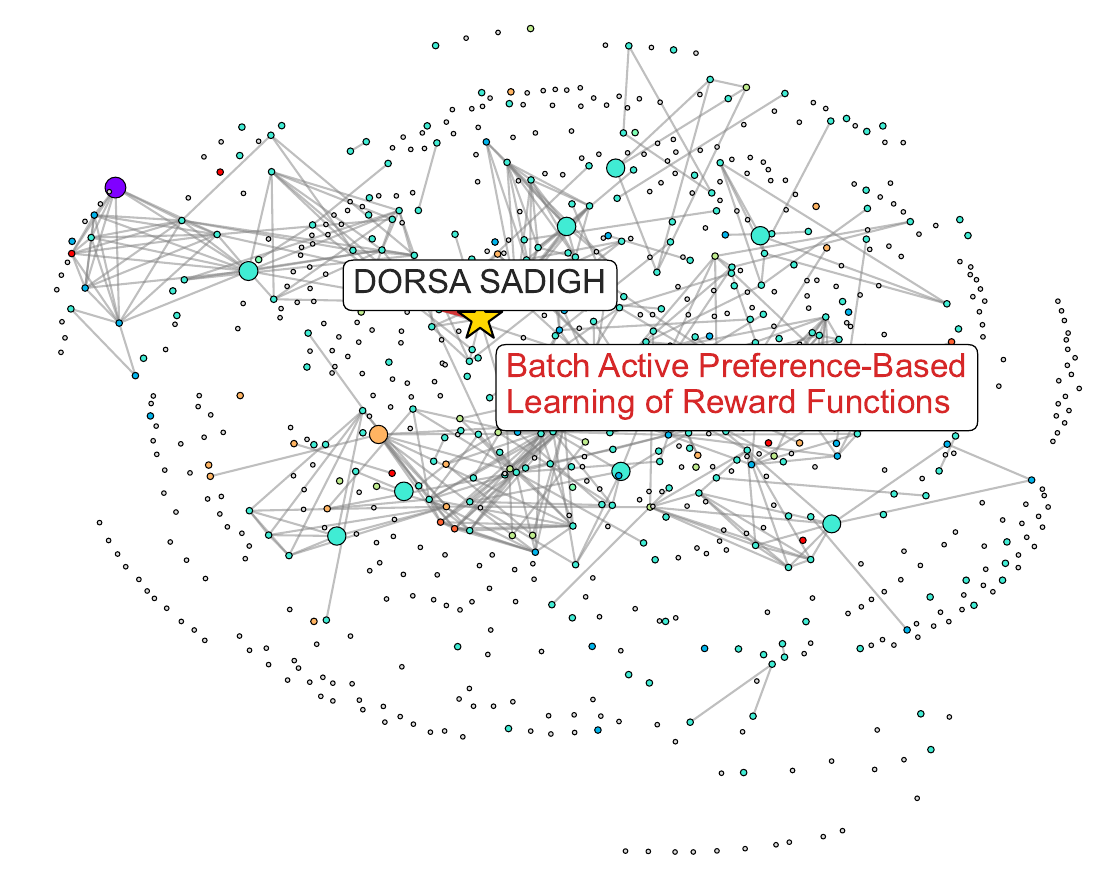}
    \end{subfigure}\hfill
    \begin{subfigure}{0.48\linewidth}
        \centering
        \includegraphics[width=\linewidth]{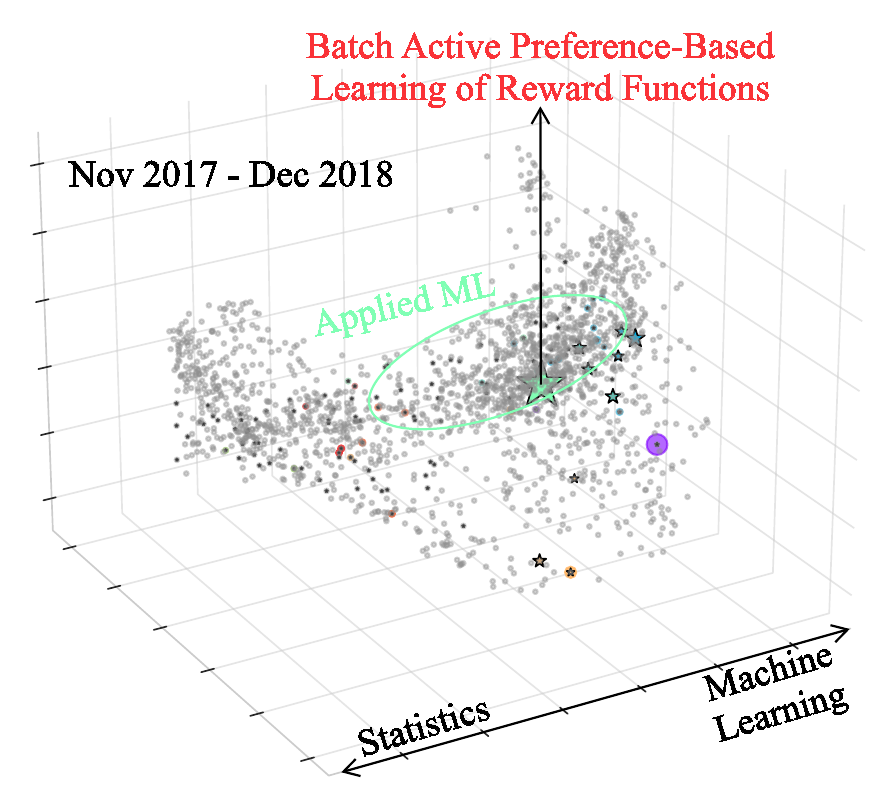}
    \end{subfigure}

    \begin{subfigure}{0.48\linewidth}
        \centering
        \includegraphics[width=\linewidth]{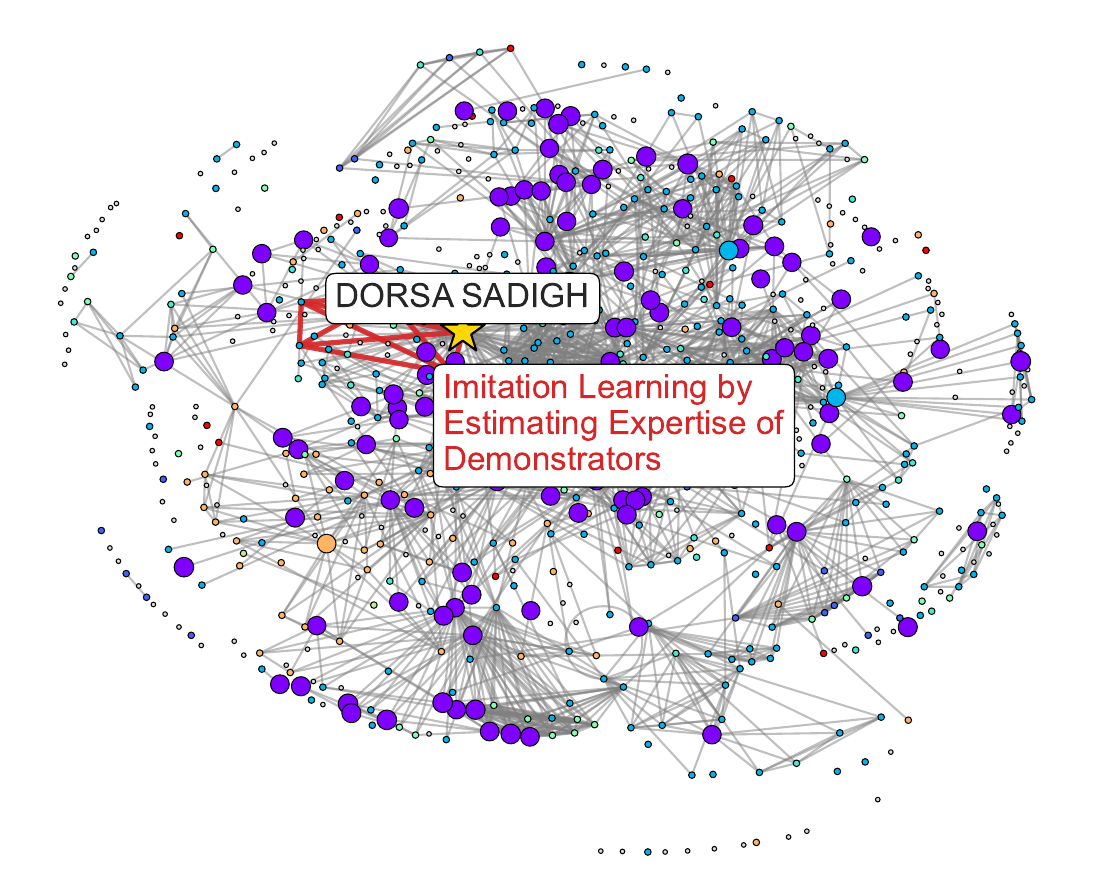}
    \end{subfigure}\hfill
    \begin{subfigure}{0.48\linewidth}
        \centering
        \includegraphics[width=\linewidth]{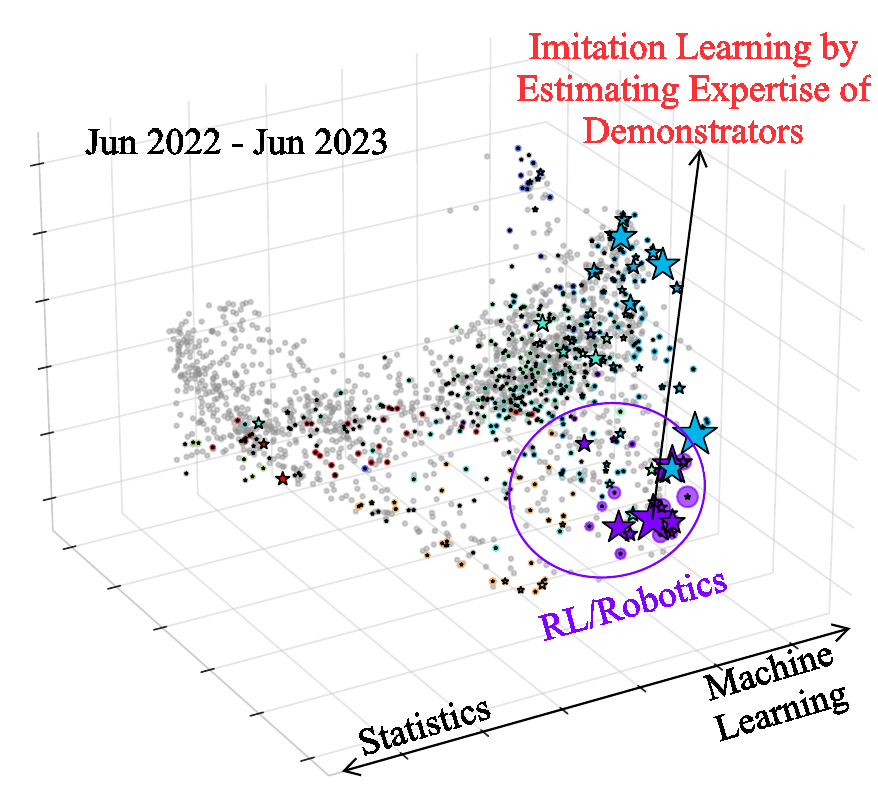}
    \end{subfigure}
    
    \caption{(Top) Hellinger-PHATE embedding of the topic manifold with 9 meta topics. 
    (Left Panels) Dorsa Sadigh is starred; network edges represent RL/Robotics community publications during 3 time windows. (Right Panels) Star sizes are proportional to Sadigh's inferred topic distributions across topic points from each time window; colors represent meta topics.}
    \label{fig:phate_and_dorsa_sadigh}
\end{figure}

The topic space forms interpretable geometric regions that can be smoothly traversed and easily annotated with word clouds (Figure \ref{fig:phate_and_dorsa_sadigh}). By mapping meta-topic distributions to co-authors, MSTML can visualize community trends by identifying topic points with high cumulative probability mass (Figure \ref{fig:phate_and_dorsa_sadigh}). This is advantageous compared to dense vector embeddings which often lack such intuitive structure. 

Figure \ref{fig:phate_and_dorsa_sadigh} shows temporal snapshots of the RL/Robotics community in both the co-author network and topic space. Network node sizes and edges reflect active documents, while colors indicate each author’s dominant topic. In the PHATE diagrams, points are scaled and colored by topic probability mass, with star markers highlighting Dorsa Sadigh’s inferred distributions. The visualizations illustrate how MSTML captures smooth temporal alignment and topic drift, with Sadigh’s path showing a shift in interests from Bayesian Methods to RL/Robotics.

\vspace{-0.2in}

\subsection{Comparisons with Transformer Models}\label{subsec:bertopic_comparison}
Table \ref{tbl:topic_coherence} compares BERTopic against CTM \cite{bianchi2021_contextualized_topic_model}, EmbedCluster (BERT with $k$-means clustering), and LDA ensemble models. LDA ensemble is MSTML in principle, but skips relevancy trimming of the term vocabulary. BERTopic achieves the highest scores on standard topic coherence metrics ($C_\text{V}$, $C_\text{UCI}$, $C_\text{NPMI}$), consistent with its use of semantically clustered sentence-transformer embeddings. 

\begin{figure}[htbp!]
    \centering
    \begin{subfigure}{0.41\linewidth}
        \centering
        \includegraphics[width=\linewidth]{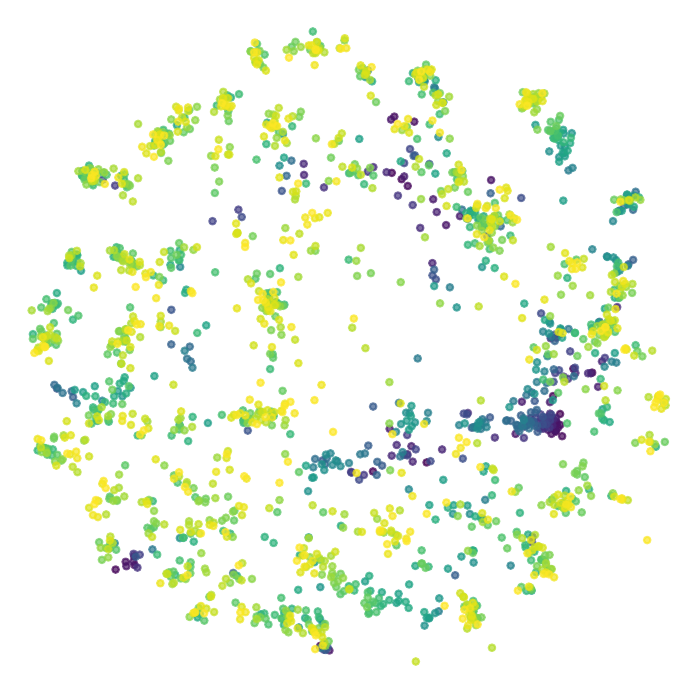}
        \caption{BERTopic (Cosine)}
        \label{fig:BERTopic_phate_t20}
    \end{subfigure}
    \hfill
    \begin{subfigure}{0.57\linewidth}
        \centering
        \includegraphics[width=\linewidth]{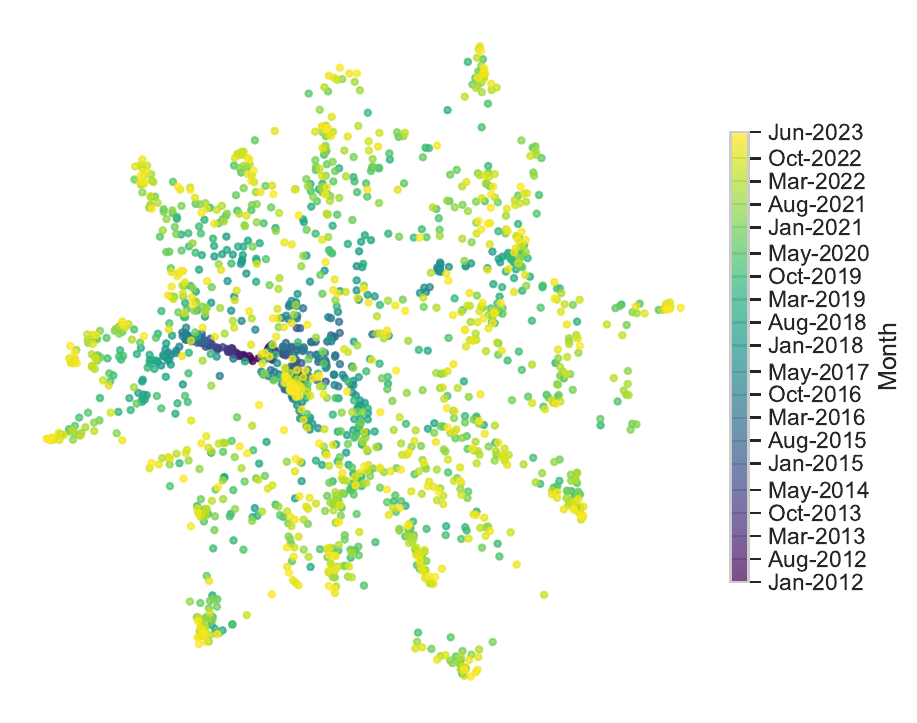}
        \caption{LDA ensemble (Hellinger)}
        \label{fig:MSTML_phate_t20}
    \end{subfigure}
    \caption{PHATE embeddings for time alignment comparison. BERTopic (left) shows tight, discrete clusters, while LDA ensemble (right) is more smooth, with clear temporal clustering.}
    \label{fig:phate_topic_alignment_compare}
\end{figure}

However, this coherence advantage comes at a cost. In terms of topic alignment, the LDA ensemble produces a continuous and temporally coherent geometry, whereas the BERTopic ensemble fragments into disconnected clusters (Figure \ref{fig:phate_topic_alignment_compare}). The temporally aligned manifold of MSTML enables tracing topic drift and detecting emerging themes, capabilities absent in models optimized primarily for semantic clustering. Quantitatively, the LDA ensemble forms a single component $k$-NN topic graph ($\text{NComp}=1$), with higher temporal neighborhood overlap (TNO) and a substantially larger spectral gap (ETSG), reflecting stronger alignment than BERTopic (Table \ref{tbl:topic_alignment}).

\begin{table}[htbp!]
\caption{Topic Coherence (Top-10 Terms)}
\vspace{-0.75em}
\centering
\label{tbl:topic_coherence}
\begin{tabular}{p{4cm}p{1cm}p{1cm}p{1cm}}
\toprule
Model Name & $C_\text{V}$ & $C_\text{UCI}$ & $C_\text{NPMI}$\\
\midrule
LDA Ensemble $\lambda=0.75$ & 0.5391 & 0.0381 & 0.0559 \\
BERTopic & \textbf{0.6081} & \textbf{0.5236} & \textbf{0.1198} \\
CTM & 0.5922 & 0.4673 & 0.0823 \\
EmbedCluster & 0.4680 & 0.2059 & 0.0316 \\
\bottomrule
\end{tabular}
\vskip -0.15in
\end{table}

\begin{table}[htbp!]
\caption{Temporal Topic Space Alignment}
\vspace{-0.75em}
\centering
\label{tbl:topic_alignment}
\begin{tabular}{p{2cm}p{1cm}p{2cm}p{2cm}}
\toprule
Model Name & $NComp$ & $TNO(w=3)$ & $ETSG(\tau=36)$\\
\midrule
LDA Ensemble & $\mathbf{1}$ & $\mathbf{0.274}$ & $\mathbf{5.188\times 10^{-3}}$\\
BERTopic & $6$ & $0.240$ & $0.219\times 10^{-3}$\\
\bottomrule
\end{tabular}
\vskip -0.3in
\end{table}

\section{Conclusion}\label{sec:conclusion}
We developed MSTML for temporal and multiscale relational topic modeling. MSTML combines probabilistic models with information geometry to build aligned and interpretable topic manifolds. Smooth topic alignment captures temporal topic drift within author and document networks. Transformer embeddings prioritize topic coherence while MSTML prioritizes topic alignment, revealing scientific novelty. Future work should explore additional ensemble-based topic models and also develop quantitative metrics for niche topic representation. Topic diversity offers a partial solution but does not fully capture the distinctiveness of niche content \cite{hougen2025mstml}. GPT-based models like GPTopic \cite{reuter2024gptopic} have been compared to MSTML but are computationally expensive \cite{hougen2025mstml}. Regardless, future work should address comparisons with GPT embeddings. We would also like to thoroughly analyze the effects of term relevancy filtering for ensemble model representations.

\newpage
\FloatBarrier
\bibliography{refs}

\end{document}